\documentclass[10pt,twocolumn,letterpaper]{article}

\usepackage[pagenumbers]{cvpr} 

\usepackage{graphicx}
\usepackage{caption}
\usepackage{amsmath}
\usepackage{amssymb}
\usepackage{booktabs}

\graphicspath{{figures/}}
\usepackage{multirow}
\usepackage{booktabs}
\usepackage{bm}
\usepackage[switch]{lineno}
\usepackage{url}
\usepackage{makecell}
\usepackage{color}
\usepackage[linesnumbered,ruled,vlined]{algorithm2e}
\usepackage{float}

\usepackage[pagebackref,breaklinks,colorlinks]{hyperref}

\usepackage[capitalize]{cleveref}
\crefname{section}{Sec.}{Secs.}
\Crefname{section}{Section}{Sections}
\Crefname{table}{Table}{Tables}
\crefname{table}{Tab.}{Tabs.}

\def\cvprPaperID{3958} 
\def\confName{CVPR}
\def\confYear{2024}

\begin{document}
	
\title{Clarity ChatGPT: An Interactive and Adaptive Processing System \\for Image Restoration and Enhancement}

\author{Yanyan Wei \and Zhao Zhang\thanks{corresponding author} \and Jiahuan Ren \and Xiaogang Xu \and Richang Hong \and Yi Yang \and Shuicheng Yan \and Meng Wang \\
}


\maketitle

\begin{abstract}
The generalization capability of existing image restoration and enhancement (IRE) methods is constrained by the limited pre-trained datasets, making it difficult to handle agnostic inputs such as different degradation levels and scenarios beyond their design scopes. Moreover, they are not equipped with interactive mechanisms to consider user preferences or feedback, and their end-to-end settings cannot provide users with more choices. Faced with the above-mentioned IRE method's limited performance and insufficient interactivity, we try to solve it from the engineering and system framework levels. Specifically, we propose Clarity ChatGPT—a transformative system that combines the conversational intelligence of ChatGPT with multiple IRE methods. Clarity ChatGPT can automatically detect image degradation types and select appropriate IRE methods to restore images, or iteratively generate satisfactory results based on user feedback. Its innovative features include a CLIP-powered detector for accurate degradation classification, no-reference image quality evaluation for performance evaluation, region-specific processing for precise enhancements, and advanced fusion techniques for optimal restoration results. Clarity ChatGPT marks a significant advancement in integrating language and vision, enhancing image-text interactions, and providing a robust, high-performance IRE solution. Our case studies demonstrate that Clarity ChatGPT effectively improves the generalization and interaction capabilities in the IRE, and also fills the gap in the low-level domain of the existing vision-language model.
\vspace{-6mm}
\end{abstract}

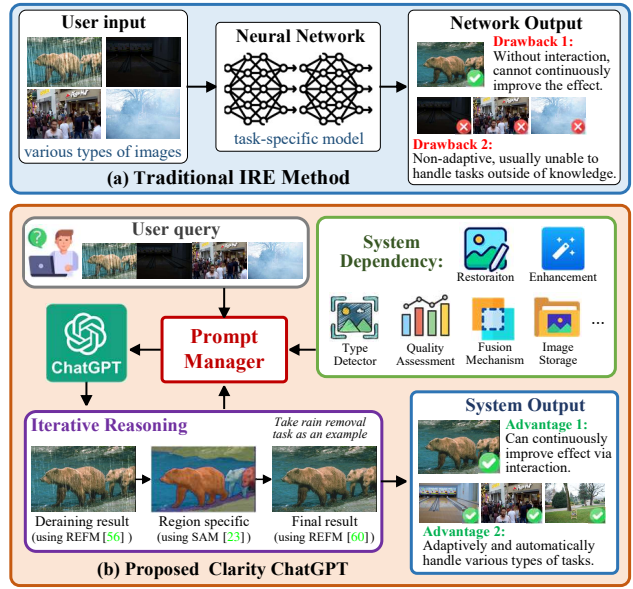
\begin{figure}[t]	
	\setlength{\unitlength}{1cm}
	\begin{center}
		\centering
        \includegraphics[width=1\linewidth]{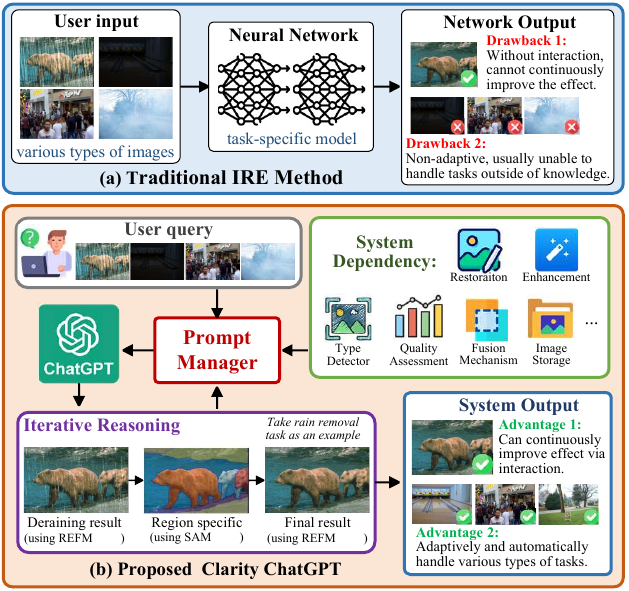}
		\begin{picture}(0,0)
        \put(-2.95,1.06){\tiny \cite{PReNet}}
        \put(-1.35,1.06){\tiny \cite{Segment-Anything}}
		\put(0.3,1.06){\tiny \cite{RCDNet}}
		\end{picture}    
	\end{center}
    \vspace{-8mm}
	\caption{This figure compares \textbf{(a)} Traditional IRE Method with \textbf{(b)} the proposed Clarity ChatGPT system. Traditional IRE methods usually rely on static neural networks to accomplish specific tasks and output non-adaptive results. In contrast, our proposed Clarity ChatGPT employs a dynamic prompt manager informed by user queries, which iteratively reasons and adapts through system dependencies like type detection and quality assessment, leading to improved outcomes through interaction and handling of a variety of tasks. Zoom in for a better view.}\label{fig:1} 
	\vspace{-6mm}
\end{figure}

\section{Introduction}
Recent years have witnessed lots of significant achievements in large language models like the GPT series \cite{GPT-2,GPT-3,ChatGPT}  and visual models like ViT \cite{VIT} in their respective domains. ChatGPT-4V \cite{GPT-4} can handle image, text and voice inputs, deeply understand image content, and even call on DALL-E 3 for image generation. However, ChatGPT-4V remains unable to fulfill user requirements for IRE tasks. At the same time, in the fundamental vision task of IRE, despite considerable progress, there are still numerous challenges to overcome: 1) \textbf{limited adaptability}: existing IRE algorithms are usually designed for specific degradation types and cannot handle unexpected variations without manual adjustments or retraining; 2) \textbf{lack of interactivity}: traditional IRE algorithms do not incorporate user feedback loops, which limits their ability to iteratively refine outputs based on user interaction. Against this background, we ask: \textit{is it possible to leverage the powerful conversational capabilities of large language models to create a system that not only integrates existing image processing technologies but also provides an intuitive and efficient user experience?}

To address this challenge, we introduce Clarity ChatGPT, an innovative system that tightly integrates large language models with advanced visual models, including Visual Foundation Models (VFMs), and Restoration and Enhancement Foundation Models (REFMs). By leveraging the capabilities of GPT-3.5 \cite{GPT-3} and specialized visual models—sourced from extensive open internet content, Clarity ChatGPT provides a direct and efficient way for users to perform complex image manipulation and enhancement via natural language interaction. The system is equipped with an automated degradation detector and no-reference image quality assessment (IQA) mechanism, which actively analyzes the input image and text, allowing for an informed and automatic selection of the most suitable models. This feature ensures that Clarity ChatGPT can intelligently translate user's text input into precise image operation instructions and call upon appropriate VFMs and REFMs to execute these operations. Consequently, users gain access to advanced IRE capabilities, eliminating the need for an in-depth understanding of image processing techniques and allowing for a dynamic response to evolving user requests and visual challenges.

Figure \ref{fig:1} provides a window into the intricate workings of Clarity ChatGPT, showcasing its adaptability and the difference from traditional IRE methodologies. Unlike traditional IRE methods that often rely on fixed models for specific tasks, Clarity ChatGPT utilizes a dynamic approach by employing a prompt manager that intelligently handles complex user queries. This system incorporates a variety of classic VFMs and EFMs, with each pre-trained on diverse datasets to address different types of image degradations. The integration of these open-source, pre-trained models within Clarity ChatGPT’s architecture allows for a flexible, comprehensive, and iterative processing workflow, which is able to optimize the performance and IRE quality of results. The rain removal example in Figure \ref{fig:1} exemplifies this by visually demonstrating the system’s capability to iteratively refine the image through successive applications of different foundation models, achieving clear and satisfactory outcomes. Thus, Clarity ChatGPT provides a more holistic, flexible, and quality-focused solution for IRE, expanding the horizons of what is achievable in the domain. The features of Clarity ChatGPT are summarized as follows:

$\bullet$ \textbf{Comprehensive Integrated System Design:} ClarityChatGPT is the first system that bridges adaptive image processing with interactive user feedback, which innovatively integrates large language and visual models. This design enables intuitive handling of IRE challenges through natural language, significantly enhancing user experience in adaptive and interactive IRE tasks.

$\bullet$ \textbf{Customized CLIP Degradation Detection:} By customizing the CLIP architecture, the system can accurately detect various types of image degradations and intelligently guide the restoration workflow, thereby improving upon the limitations of conventional IRE methods and allowing for smarter and more efficient processing.

$\bullet$ \textbf{Instant No-Reference Evaluation Mechanism:} The integration of no-reference IQA models offers ordinary users immediate evaluation of image processing outcomes, a feature traditionally limited to expert use and scholarly settings.

$\bullet$ \textbf{Region-Specific Optimization Strategy:} Based on utilizing state-of-the-art image segmentation and object detection technologies, ClarityChatGPT achieves meticulous treatment of specific image regions, offering local optimization capabilities that are not found in traditional tools.

$\bullet$ \textbf{Innovative Multiple Results Fusion Technology:} The system uses a novel fusion method to cohesively blend results from different processing techniques, ensuring visual consistency and superior quality in final output, thereby advancing past the constraints of traditional IRE approaches.

\begin{figure*}[t]
    \centering
    \begin{minipage}{0.65\textwidth}
        \centering
        \includegraphics[width=\textwidth]{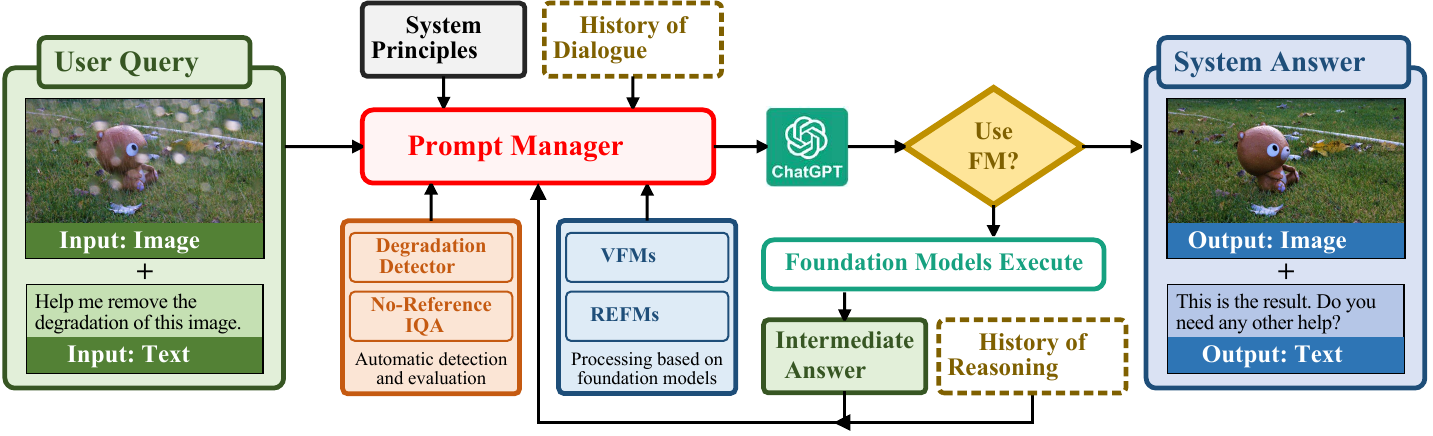}
        \begin{picture}(0,0)
            \put(-114,95){\tiny $\mathcal{Q}_{i}$}
            \put(-113,54){\tiny \textcolor{white}{$I_i$}}
            \put(-115,29){\tiny \textcolor{white}{$T_i$}}
            \put(-50,97){\tiny $\mathcal{P}$}
            \put(-13,97){\tiny $\mathcal{H}_{<i}$}
            \put(-17,75){\scriptsize \textcolor{red}{$\mathcal{M}$}}
            \put(-56,48){\tiny $\mathcal{D}$}
            \put(-58,35){\tiny $\mathcal{E}$}
            \put(-10,51){\tiny $\mathcal{F}_V$}
            \put(-10,38){\tiny $\mathcal{F}_D$}
            \put(36,25){\tiny $\mathcal{A}_i^{j}$}
            \put(79,25){\tiny $\mathcal{R}_i^{<j}$}
            \put(149,95){\tiny $\mathcal{A}_{i}$}
            \put(145,54){\tiny \textcolor{white}{$I'_i$}}
            \put(145,29){\tiny \textcolor{white}{$T'_i$}}
        \end{picture}    
        \vspace{-4mm}
        \caption{Clarity ChatGPT processes user-submitted images and text through degradation detection, foundation model execution, and output generation with interactive feedback.}
        \label{fig:2}
        \vspace{-4mm}
    \end{minipage}
    \hfill
    \begin{minipage}{0.3\textwidth}
        \centering
        \resizebox{\linewidth}{!}{
        \begin{tabular}{c l c}
            \toprule[1pt]
            ~ & \textbf{Task} & \textbf{Model} \\	
            \midrule[0.5pt]
            ~ & Segmentation & SAM \cite{Segment-Anything} \\
            \multirow{1}{*}{\textbf{VFMs}} & Detection & GroundingDINO \cite{Detection-Anything} \\
            ~ & Caption & BLIP \cite{I2T} \\
            \midrule[0.5pt]
            ~ & Derain & PReNet \cite{PReNet}, JORDER \cite{JORDER}, RCDNet \cite{RCDNet} \\
            ~ & Dedrop & attentive GAN \cite{attentiveGAN}, DPRRN \cite{DPRRN}, He \textit{et.al} \cite{raindrop3}\\
            ~ & Desnow & DesnowNet \cite{DesnowNet}, HDCWNet \cite{HDCWNet}, SnowFormer \cite{Snowformer} \\
            \multirow{8}{*}{\textbf{REFMs}} & Denoise & DnCNN \cite{DnCNN}, VDN \cite{VDN}, Neighbor2Neighbor \cite{Neighbor2Neighbor} \\
            ~ & Deblur & SRN-DeblurNet \cite{SRN-DeblurNet}, DeepDeblur \cite{DeepDeblur}, DeblurGAN \cite{Deblurgan} \\
            ~ & Dehaze & FFA-Net \cite{FFA-Net}, GridDehazeNet \cite{Griddehazenet}, DehazeNet \cite{DehazeNet} \\
            ~ & Dewatermark & WDNet \cite{Watermark1}, Niu \textit{et.al} \cite{Watermark2}, Cun \textit{et.al} \cite{Watermark3} \\
            ~ & Deshadow & Le \textit{et.al} \cite{Shadow1}, G2R-ShadowNet \cite{Shadow2}, CANet \cite{Shadow3} \\
            ~ & Deflare & Zhang \textit{et.al} \cite{Flare3} Wu \textit{et.al} \cite{Flare1}, Qiao \textit{et.al} \cite{Flare2} \\
            ~ & Overexposure & ENCNet \cite{ENCNet}, ERL \cite{ERL}, Eyiokur \textit{et.al} \cite{Exposure3} \\
            ~ & Inpaint & Yu \textit{et.al} \cite{inpaint1}, RePaint \cite{RePaint}, MAT \cite{MAT} \\
            ~ & JPEG & FBCNN \cite{FBCNN}, QGCN \cite{QGCN}, Liu \textit{et.al} \cite{JPEG3} \\
            ~ & LLE & DCC-Net \cite{DCC-Net}, SNR-Net \cite{SNR}, LLFlow \cite{LLFlow} \\
            ~ & ALL-in-one & IPT \cite{IPT}, TransWeather\cite{TransWeather1}, CPNet \cite{BID} \\
            \bottomrule[1pt]
        \end{tabular}
        }
        \captionof{table}{Foundation models (VFMs and REFMs) supported by Clarity ChatGPT.}
        \label{table:1}
        \vspace{-4mm}
    \end{minipage}
\end{figure*}

\section{Related Work}\label{Related}
\subsection{Image Restoration and Enhancement (IRE)}
IRE are two central branches of digital image processing, with the primary goal of optimizing visual quality. Image restoration focuses on rectifying distortions and degradations in images due to factors such as noise, blur, camera misalignment, motion blur, and atmospheric scattering. This involves tasks like denoising \cite{DnCNN,VDN}, deblurring \cite{DeepDeblur,Deblurgan}, deraining \cite{JORDER,PReNet,RCDNet,Deraincyclegan}, and dehazing \cite{FFA-Net,Griddehazenet}. In contrast, image enhancement emphasizes improving the perceptual quality of images and refining certain visual effects. This encompasses tasks such as super-resolution \cite{SRCNN,LapSRN,TTSR}, low-light enhancement \cite{Retinex,ZeroDCE,DCC-Net}, flare removal \cite{Flare1,Flare2,Flare3}, shadow removal \cite{Shadow1,Shadow2,Shadow3}, watermark removal \cite{Watermark1,Watermark2,Watermark3}, and overexposure correction \cite{ENCNet,ERL,Exposure3}. These techniques not only accentuate or refine certain features of images but also improve their overall visual appeal.

In the IRE domain, researchers usually focus on specific tasks, such as denoising, deblurring, or super-resolution. However, these specialized models often face adaptability challenges in real-world scenarios. As a result, some researchers have begun exploring models that can broadly address multiple tasks \cite{Restormer,Swinir,IPT,BID}, even though this might entail integrating several pre-trained models. These challenges have spurred further research aimed at enhancing the robustness and generalization capabilities of models, striving to develop solutions that can simultaneously handle various image degradation and enhancement tasks \cite{TransWeather1,TransWeather2,TransWeather3,TransWeather4}. Yet, due to significant disparities between tasks and models, deploying them in the real world remains challenging. Currently, there's an urgent need in the research community for a truly universal IRE system that can meet a wide range of practical application demands. To change this status quo, Clarity ChatGPT is introduced as an innovative solution, which merges the conversational capabilities of Large Language Models (LLMs) with the processing strengths of existing IRE algorithms. By integrating fine-grained processing and multi-result fusion strategies, it aims to provide users with a more universal and intuitive IRE interface, addressing the robustness and generalization demands of IRE tasks in real-world scenarios.

\subsection{ChatGPT-based Visual Agent}
Chain-of-Thought (CoT) \cite{COT1,COT2,COT3,COT4} is a specially designed technique aimed at maximizing the multi-step reasoning capabilities of LLMs. Unlike traditional methods that seek direct answers from models, the CoT strategy demands the generation of intermediate answers, bridging the gap between the question and the final response. This stepwise reasoning mirrors human thought processes, allowing for a more intricate, detailed, and potentially accurate interaction. Against this backdrop of reasoning, the ChatGPT-based Agent emerges, representing an innovative fusion of conversational AI and visual processing prowess. It not only inherits the foundational strengths of ChatGPT in natural language understanding and generation but also extends further into visual tasks, ensuring deep and accurate interactions with users. In essence, this integration opens up a novel research direction, aiming to extend the potential of stepwise reasoning across tasks, from text-to-image generation to image-to-text transformations. Systems like Visual ChatGPT \cite{Visual-Chatgpt}\footnote{https://github.com/microsoft/TaskMatrix} and Video ChatGPT \cite{Video-ChatGPT}\footnote{https://github.com/mbzuai-oryx/Video-ChatGPT} epitomize this integration, allowing users to not only engage in natural, fluid dialogues but also to instruct, query, and receive feedback on visual content. Such a synthesis offers a more interactive and intuitive way for users to interface with AI systems, bridging the gap between textual conversation and visual comprehension, and paving the way for myriad applications across various domains. In addition to the aforementioned, many researchers are exploring the use of ChatGPT for dialogic applications in more specialized and vertical domains. For instance, Qin \textit{et al}. \cite{qin2023read} delves into combining ChatGPT with large language models, offering an explainable and interactive approach to depression detection on social media platforms. Liu \textit{et al}. \cite{liu2023chatgpt} investigates the potential of leveraging ChatGPT to establish an explainable framework for zero-shot medical image diagnostic procedures. These explorations underscore the extensive potential of ChatGPT in professional domains such as mental health screening and medical image analysis, showcasing its unique value in various intricate scenarios.

\section{Proposed Clarity ChatGPT}\label{Method}
Figure \ref{fig:2} illustrates the pipeline of our Clarity ChatGPT. Consider a dialogue system $\mathcal{S}$, consisting of \textit{N} question-answer pairs: $\mathcal{S} = \{(\mathcal{Q}_1, \mathcal{A}_1),(\mathcal{Q}_2, \mathcal{A}_2), ...,(\mathcal{Q}_N, \mathcal{A}_N )\}$. In the \textit{i}-th round of dialogue, based on user's query $\mathcal{Q}_i$, to derive the response $\mathcal{A}_i$, the system employs multiple visual foundation models (VFMs), restoration and enhancement foundation models (REFMs) along with their respective intermediate outputs $\mathcal{A}_i^{(j)}$. Here, \textit{j} signifies the output stemming from the \textit{j}-th VFM and REFM during the \textit{i}-th round. Based on the above basic definition, we can first propose the formal definition of Clarity ChatGPT:
\begin{equation} \label{eqn1}
    \footnotesize
    \begin{aligned}
    \mathcal{A}_i^{(j+1)}=&ChatGPT(\mathcal{M}\left ( \mathcal{P}\right ),\mathcal{M}\left ( \mathcal{D}\right ),\mathcal{M}\left ( \mathcal{E}\right ),\mathcal{M}\left ( \mathcal{F}_V, \mathcal{F}_D\right ),\\
    &\mathcal{M}\left ( \mathcal{H}_{<i}\right ),
    \mathcal{M}\left ( \mathcal{Q}_{i}\right ), \mathcal{M}(\mathcal{R}_i^{(<j)}),\mathcal{M}(\mathcal{F}(\mathcal{A}_i^{(j)}))),
    \end{aligned}
\end{equation}

\noindent where $\mathcal{M}$ is the prompt manager used to transform all visual inputs into a linguistic format, making them comprehensible to the ChatGPT model. $\mathcal{P}$ is the system principle for Clarity ChatGPT, including considerations for image filenames and the preference for using VFMs and REFMs for image tasks over chat history. $\mathcal{D}$ is the degradation detector, used for automatic degradation detection. $\mathcal{E}$ is the no-reference IQA metrics, utilized for evaluating image quality. $\mathcal{F}_V$ and $\mathcal{F}_D$ represent VFMs and REFMs, used for processing images (see Table \ref{table:1} for details). $\mathcal{H}_{<i}$ denotes all historical information before the \textit{i}-th round. $\mathcal{Q}_{i}$ is the user query in the \textit{i}-th round. $\mathcal{R}_i^{(<j)}$ represents the accumulated reasoning histories derived from the \textit{j} previously engaged REFMs during the \textit{i}-th conversation cycle. $\mathcal{A}_i^{(j)}$ signifies the output of the \textit{j}-th VFM and REFM in the \textit{i}-th round.

\begin{figure}[t]
	\centering
	\includegraphics[width=\linewidth]{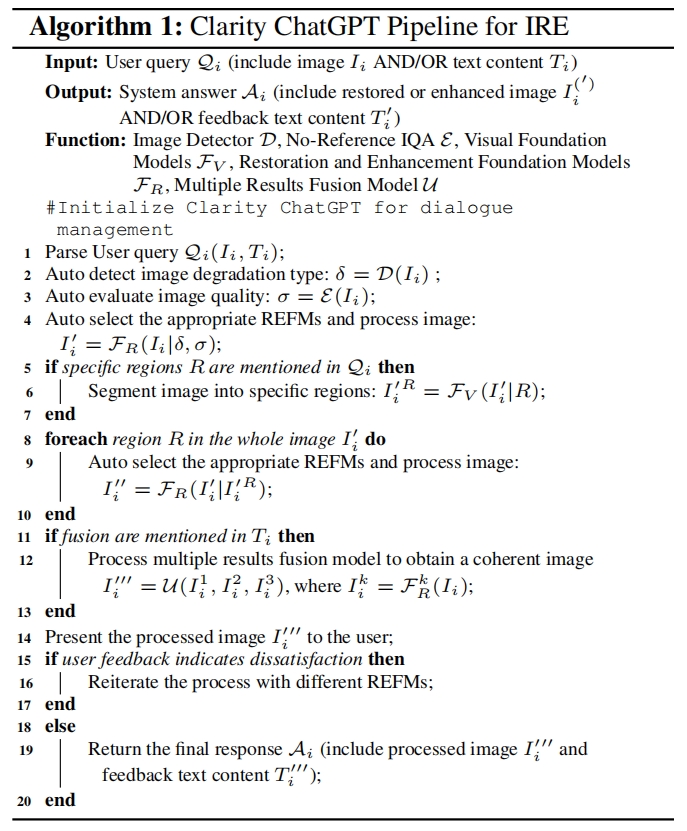}
	\label{fig:a}
	\vspace{-4mm}
\end{figure}

We show the process of Clarity ChatGPT in Algorithm 1. The system first obtains the input image $I_i$ and parses the user's textual query $T_i$; then it uses the Image Detector $\mathcal{D}$ to detect the degradation type $\delta$ of the input image and employs the No-Reference IQA $\mathcal{E}$ to evaluate the image quality $\sigma$. Subsequently, it selects the appropriate REFMs based on $\delta$ and $\sigma$ and processes the image. If specific image regions $R$ are mentioned in the user’s query $\mathcal{Q}_i$, the system performs VFMs $\mathcal{F}_V$ on image $I'_i$ to obtain segmentation image $I_i^{\prime R}$; then the system can auto-select the appropriate REFMs to obtain restored image $I''_i$. Additionally, if multiple results fusion demand is required, the system applies fusion model $\mathcal{U}$ to obtain a coherent image $I'''_i=\mathcal{U}(I_i^1, I_i^2, I_i^3)$ based on multiple results of REFMs. Throughout the process, the system generates intermediate responses $\mathcal{A}_i^{(j)}$ and presents them to the user. If the user is not satisfied with the results, the system will repeat the process with different REFMs. Ultimately, the system generates $\mathcal{A}_i$ as the final response, terminating any further execution of VFMs and REFMs.

\subsection{Prompt Managing of Degradation Detector}
The integration of the CLIP \cite{CLIP} network within Clarity ChatGPT serves as a cornerstone for our prompt management strategy. Designed to understand and categorize images in concert with natural language descriptions, CLIP's zero-shot learning capabilities are harnessed and enhanced in our system to address the intricate challenges of image degradation type detection. In traditional settings, CLIP's pre-trained model, although powerful, fell short in terms of precise degradation type identification due to the generalized nature of its training data. Within the Clarity ChatGPT framework, we have refined the CLIP model's comprehension of image degradation through targeted finetuning on a specially curated dataset. This dataset is tailored to represent a wide array of degradation scenarios, enumerated as `normal', `rain streak', `raindrop', `snow', and twelve other distinct conditions. The original CLIP model achieved only a Top-1 accuracy of 38.27\% while improving the Top-1 accuracy to 94.57\% after fine-tuning (see Table \ref{table:2} for details).

This customization enables the prompt management system to discern between different degradation types with enhanced accuracy, feeding into a more coherent and effective image restoration pipeline. The Prompt Manager, by leveraging this tailored detection capability, is thereby empowered to guide the subsequent processing stages in a more informed and dynamic manner. This leads to a versatile and adaptable system, capable of handling various image enhancement tasks with improved proficiency. The positive implications of this enhancement are evidenced not only by the empirical results shown in the experimental section but also by the system's increased agility in adjusting to a diverse set of image degradation challenges.

\begin{figure}[t]
	\centering
	\includegraphics[width=\linewidth]{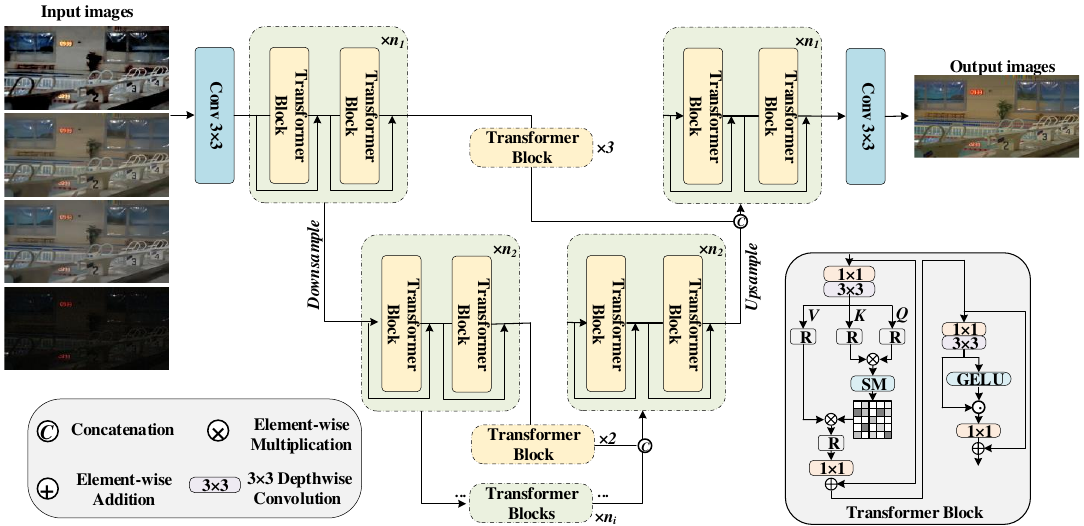}
	\caption{The structure of proposed fusion network.}
	\label{fig:3}
    \vspace{-4mm}
\end{figure}

\subsection{Prompt Managing of No-Reference IQA}
In the image processing workflow, we underscore the pivotal role of no-reference IQA. Our system integrates multiple no-reference IQA metrics such as SSEQ \cite{SSEQ}, NIQE \cite{NIQE}, and BRISQUE \cite{BRISQUE}, enabling automatic and instant quality scoring following each image input or output. This immediate quality feedback mechanism provides users with an intuitive standard for evaluating image quality and identifying potential visual imperfections. Moreover, acknowledging the distinct applicability of various metrics to different scenarios, we implement a weighted scoring approach to holistically reflect image quality. This automated assessment process not only streamlines the traditional procedure reliant on reference images but also enhances the system's real-time responsiveness and user interactivity. In practical applications, this autonomous NR-IQA offers users a swift and reliable tool, boosting their confidence in decision-making and catering to their specific needs and expectations more aptly.

\begin{figure}[t]
	\centering
	\includegraphics[width=\linewidth]{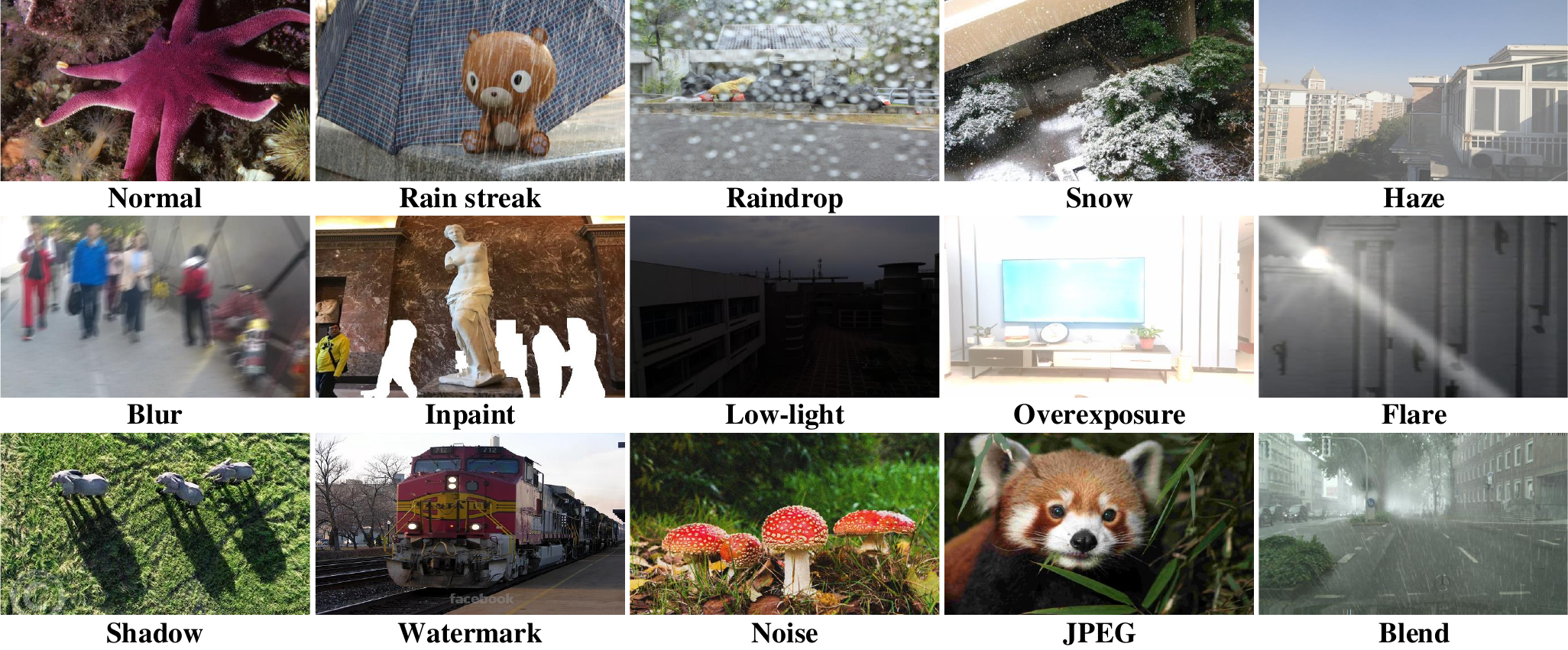}
    \vspace{-6mm}
	\caption{Examples of 15 different types of images in the proposed dataset for fine-tuning the CLIP \cite{CLIP} model.}
	\label{fig:4}
    \vspace{-4mm}
\end{figure}

\subsection{Prompt Managing of Regional Improvement}
Traditional IRE methods typically adopt an end-to-end approach, which is limited to whole processing process and fails to focus on specific image areas. A significant limitation of these methods is the absence of interactive visual functionalities, preventing users from directing the model to process designated regions with precision. Furthermore, traditional methods, focusing solely on overall IRE, do not support segmentation or object detection algorithms, thus failing to meet user-specific demands on a granular level.

To overcome these issues, we introduce a region-specific refinement technique, which combines the SAM \cite{Segment-Anything} for precise object segmentation with GroundingDINO \cite{GroundingDINO} for accurate object detection, opening new avenues for IRE tasks. Users can now specify areas of interest for processing, which are precisely delineated using aforementioned methods. Subsequently, specialized IRE models are applied solely to these designated areas, achieving targeted optimization. This strategy ensures that only segments of interest to the user are enhanced, meeting user needs in detail, and increasing processing precision and efficiency, thereby improving user satisfaction and interactive experience.

\subsection{Prompt Managing of Multiple Results Fusion}
Note the fact that different image restoration methods usually produce varying levels of quality when applied to degraded images. To address this issue and make the obtained restored images contain more detailed information, we propose to design a multi-result fusion network to improve the quality of restored images with multi-inputs. 

\begin{figure}[t]
    \centering
    \begin{minipage}{0.25\textwidth}
        \centering
    	\includegraphics[width=\linewidth]{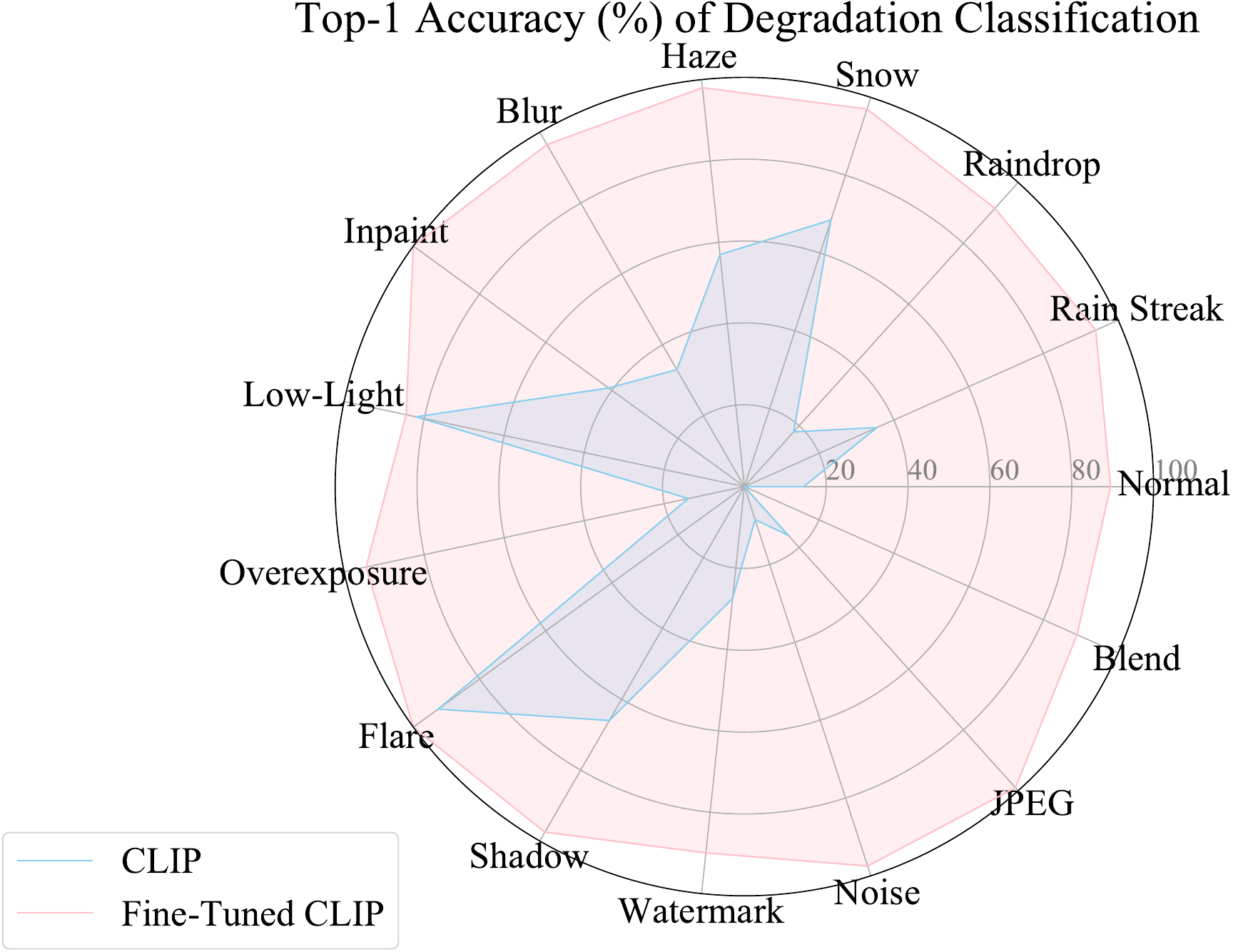}
        \vspace{-6mm}
    	\caption{Top-1 degradation classification accuracy (\%) of original CLIP \cite{CLIP} and fine-tuned CLIP.}
        \label{fig:5}
        \vspace{-6mm}
    \end{minipage}\hfill
    \begin{minipage}{0.2\textwidth}
        \centering
        \resizebox{\linewidth}{!}{
        \begin{tabular}{l c c}
            \toprule[1pt]
            \textbf{Accuracy} & CLIP \cite{CLIP} & CLIP (finetuned) \\
            \midrule[0.5pt]
            Top-1 & 38.27\% & \textbf{94.57\%} \\
            Top-3 & 60.00\% & \textbf{99.67\%} \\
            Top-5 & 71.20\% & \textbf{99.87\%} \\
            \bottomrule[1pt]
        \end{tabular}
        }
        \captionof{table}{Top-\textit{k} (i.e., 1, 3, 5) accuracy (\%) for the degradation classification by the original CLIP \cite{CLIP} and fine-tuned CLIP.}
        \label{table:2}
        \vspace{-6mm}
    \end{minipage}
\end{figure}

This approach aligns with the concept of leveraging multiple restoration results of three different methods to improve image quality. By combining the outputs of different restoration methods, it is possible to harness the strengths of each method and mitigate their individual limitations. This fusion process aims to generate a final restored image that contains enhanced details and exhibits improved overall quality compared to using a single restoration method. We take the low-light image enhancement as an example to describe the fusion process.

Specifically, we take the enhanced results of three LLIE methods, i.e., DCC-Net \cite{DCC-Net}, SNR \cite{SNR}, and LLFlow \cite{LLFlow} and the original low-light images as input for further processing or analysis. The input tensor can be formulated as $F_{input}=concatenate(F_{DCC}, F_{SNR}, F_{LLFlow}, F_{Low})\in\mathbb{R}^{12\times h\times w}$, where $F_{DCC}$, $F_{SNR}$ and $F_{LLFlow}$ are the restored images of DCC-Net, SNR, and LLFlow respectively. $F_{Low}$ is the low-light image. By incorporating the enhanced results from these LLIE methods, we aim to leverage their respective strengths and improve the overall quality and visual characteristics of the low-light images.  
Inspired by efficient transformers, our fusion network is designed based on the transformer block introduced in the Restormer~\cite{zamir2022restormer}. The overall framework of our fusion network is depicted in Figure \ref{fig:3}, which adopts a UNet structure. It comprises three encoding blocks and three decoding blocks. 
Specifically, each encoder has four transformer blocks to extract hierarchical features with increasing levels of abstraction. The transformer blocks within each encoder contribute to capturing the spatial and contextual information in the input. Similar to the encoder, each decoder block has four transformer blocks and aims to reconstruct the high-resolution output. To further enhance the integration of features across different levels, we adopt a specific configuration for the transformer blocks in the skip connections. We set up three, two, and one transformer blocks, respectively, in different levels of skip connections. This arrangement ensures that features from multiple levels are effectively integrated and utilized during the fusion process. The process of proposed fusion network can be formulated as
\begin{equation}
    \small
	\begin{split}
		&F_{shallow} = f_{3 \times 3}(F_{input})\\
            &F_{encoder} = Ds(Ed_3(Ds(Ed_2(Ds(Ed_1(F_{shallow}))))))\\
            &F_{latent} = \underbrace{Transformer(\cdots Transformer(F_{encoder}))}_{n_{4}}\\
            &F_{decoder} = Us(Dc_1(Us(Dc_2(Us(Dc_3(F_{latent}))))))\\
            &F_{output} = f_{3 \times 3}(F_{decoder})
	\end{split}
	\label{eq:3}
\end{equation}
where $f_{3 \times 3}(\cdotp)$ is a $ 3 \times 3$ convolution. $Ed_i(\cdotp)$ and $Dc_i(\cdotp)$ denote the encoder and decoder, respectively. $Ds(\cdotp)$ and $Us(\cdotp)$ are the downsample and upsample operations.

\begin{figure}[t]
	\centering
	\includegraphics[width=\linewidth]{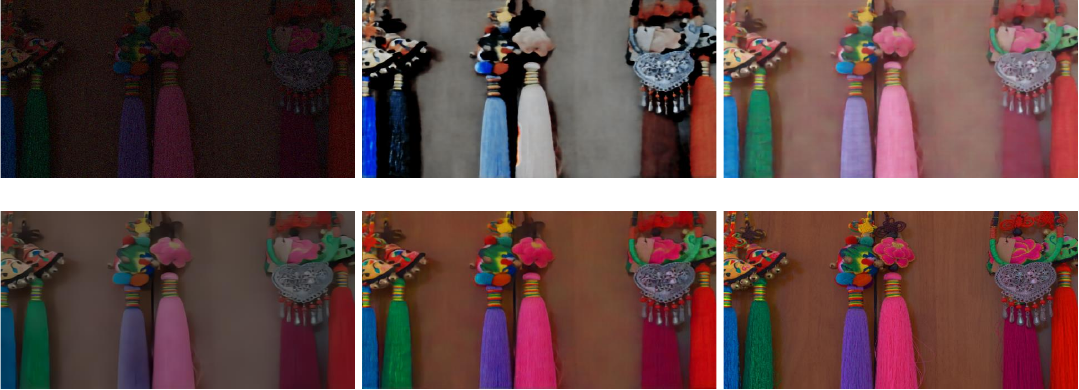}
	\begin{picture}(0,0)
		\put(-85,53){\scriptsize Input}
        \put(-95,60){\scriptsize \textcolor{white}{12.33/0.243}}
		\put(-19,53){\scriptsize DCC-Net \cite{DCC-Net}}
        \put(-16,60){\scriptsize \textcolor{white}{14.30/0.541}}
		\put(61,53){\scriptsize SNR-Net \cite{SNR}}
        \put(64,60){\scriptsize \textcolor{white}{15.37/0.664}}
		\put(-96,6){\scriptsize LLFlow \cite{LLFlow}}
        \put(-95,13){\scriptsize \textcolor{white}{19.39/0.735}}
        \put(-4,6){\scriptsize Ours}
        \put(-15,13){\scriptsize \textcolor{white}{28.36/0.832}}
        \put(63,6){\scriptsize Ground truth}
        \put(64,13){\scriptsize \textcolor{white}{$+\infty$/1.000}}
	\end{picture}  
    \vspace{-4mm}
	\caption{Comparison of the visualization results of the proposed fusion strategy and other methods on the low-light+noise dataset. The white font is the PSNR and SSIM indicators of the image.}
    \vspace{-4mm}
	\label{fig:6}
\end{figure}

\section{Experiments}\label{Experiments}
\subsection{Setup}
We implement the LLM with ChatGPT \cite{ChatGPT} (OpenAI ``text-davinci-003" version), and guide the LLM with LangChain \cite{LangChain}\footnote{https://github.com/langchain-ai/langchain}. We collect restoration and enhancement foundation models (as well as some familiar visual foundation models, such as segmentation and detection) from open-source platforms, such as GitHub\footnote{https://github.com/} and HuggingFace\footnote{https://huggingface.co/} (see Table \ref{table:1} for details). Since the sizes of restored and enhanced models vary, we recommend users flexibly load VFM and REFM as needed. Depending on the task, the required GPU memory size may vary from 12Gb to 48Gb. The maximum length of chat history is 2,000, and excessive tokens are truncated to meet the input length of ChatGPT.

\subsection{Performance of Degradation Detection}
We constructed a diverse dataset with 15 degradation types, including `normal’, `rain streak', `raindrop', `snow', `haze', `blur', `inpaint', `low-light', `overexposure', `flare', `shadow', `watermark', `noise', `JPEG' and `blend'. Each category has 1,000 images, with 800 images allocated for training and 200 for testing. These images were sourced from reputable synthetic and real-world datasets, with a collection ratio of synthetic to real being 7:3. Figure \ref{fig:4} shows examples of each type. The source of specific data can be found in the \textit{supplementary material}. The used assessment metric is the standard classification accuracy used to gauge the performance of the degradation detector.

The experimental analysis presented in Figure \ref{fig:5} and Table \ref{table:2} reveals a stark disparity in accuracy between the original CLIP model and its fine-tuned counterpart across various types of image degradation. Initially, the CLIP model's average accuracy lingered at 38.27\%, which pointed to substantial difficulties in handling the diverse image quality challenges. The model particularly struggled with `normal' (14.5\%), `overexposure' (14\%), `noise' (8.5\%), and `blend' (0\%) categories, reflecting its initial limitation in the precise identification of these conditions. However, subsequent fine-tuning has propelled the CLIP model to remarkable accuracy heights, with an average of 94.57\%. Post-fine-tuning performance soared in all categories, with `normal' and `overexposure' detection improving to 89.5\% and 94.5\%, respectively, and `noise' and `blend' accuracy elevating dramatically to 97.5\% and 89\%. Perfect scores achieved in `inpaint' and `flare' categories illustrate the model's refined ability to tackle complex classification tasks with impeccable precision. The comprehensive performance boost demonstrates the fine-tuned CLIP's potential in delivering highly accurate and reliable image degradation classification, affirming the effectiveness of model refinement for specialized detection applications.

\begin{figure*}[t]	
	\setlength{\unitlength}{1cm}
	\begin{center}
		\centering  
        \includegraphics[width=\linewidth]{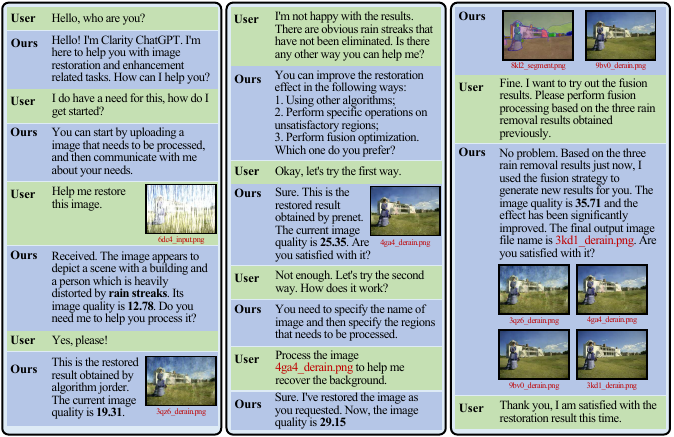}	
		\begin{picture}(0,0)
		\end{picture}    
	\end{center}
    \vspace{-8mm}
	\caption{\mdseries Multiple rounds of dialogue between humans and Clarity ChatGPT. In the dialogues, Clarity ChatGPT can understand human intents, support the language and image inputs, and accomplish IRE tasks, as well as other familiar visual tasks, such as segmentation.}\label{fig:7} 
	\vspace{-4mm}
\end{figure*}

\begin{table}[t]
    \centering
    \scriptsize
    \vspace{-2mm}
    \resizebox{\linewidth}{!}{
    \begin{tabular}{lccccc}
    \toprule[1pt]
    \textbf{Metrics} & DCC-Net \cite{DCC-Net} & SNR-Net \cite{SNR} & LLFlow \cite{LLFlow} & Ours & \textit{Improvement} \\ 
    \midrule[0.5pt]
    PSNR        & 15.35   & 21.72     & 22.28   & \textbf{27.23} & \textit{22.2\%} \\
    SSIM             & 0.601    & 0.772      & 0.784    & \textbf{0.823} & \textit{4.97\%} \\ 
    \bottomrule[1pt]
    \end{tabular}
    }
    \vspace{-2mm}
    \caption{Quantitative results on the LOL \cite{Retinex} with noise level=10.}
    \label{table:3}
    \vspace{-4mm}
\end{table}

\subsection{Multiple Results Fusion}
In the experimental assessment of our multiple results fusion module, we conducted tests on the task of low-light image enhancement combined with denoising as an example. We selected the LOL dataset \cite{Retinex}, which contains low-light images with an added noise level of 10, as the basis for evaluation. The performance of the fusion strategy was benchmarked against three state-of-the-art methods: DCC-Net \cite{DCC-Net}, SNR-Net \cite{SNR} and LLFlow \cite{LLFlow}. The quantitative results are shown in Table \ref{table:3}, which indicated that our fusion approach significantly improved the restoration quality, yielding a higher PSNR \cite{PSNR}/SSIM \cite{SSIM} score of 27.23/0.82 and a significant PSNR improvement of 22.2\%. To provide a qualitative perspective on the fusion module's effectiveness, we selected representative images and displayed them in Figure \ref{fig:6}, which illustrates the direct comparison between the input images with low-light and noise impairments and the restored outputs using the mentioned methods and our fusion-based solution. The PSNR and SSIM scores are annotated on each image, providing an apparent reference for the enhancement achieved. Specifically, the figure denotes that our method outperforms the individual techniques with a PSNR of 28.36 and an SSIM of 0.832, which is a considerable improvement over the scores from the mentioned methods' outputs. The depicted images corroborate our hypothesis that fusing the outputs of different algorithms results in superior detail retrieval and overall image quality. This result demonstrates the potential of our fusion strategy in approaching real-world image fidelity and is expected to address the significant challenges of detail loss and texture corruption present in IRE tasks. For more fusion results on different tasks, please refer to the \textit{supplementary material}.

\subsection{A Full Case of Multiple Rounds Dialogue}
Figure \ref{fig:7} illustrates a comprehensive 16-round multi-modal dialogue case, showcasing the versatility and robustness of Clarity ChatGPT. In this intricate exchange, the user engages in a dynamic conversation, posing queries that encompass both textual and visual elements. Clarity ChatGPT, in turn, responds adeptly with a combination of text and images, demonstrating its capacity for nuanced multi-modal interaction.

\subsection{Case Study of Clarity ChatGPT}
\begin{figure*}[t]
	\centering
	\includegraphics[width=\linewidth]{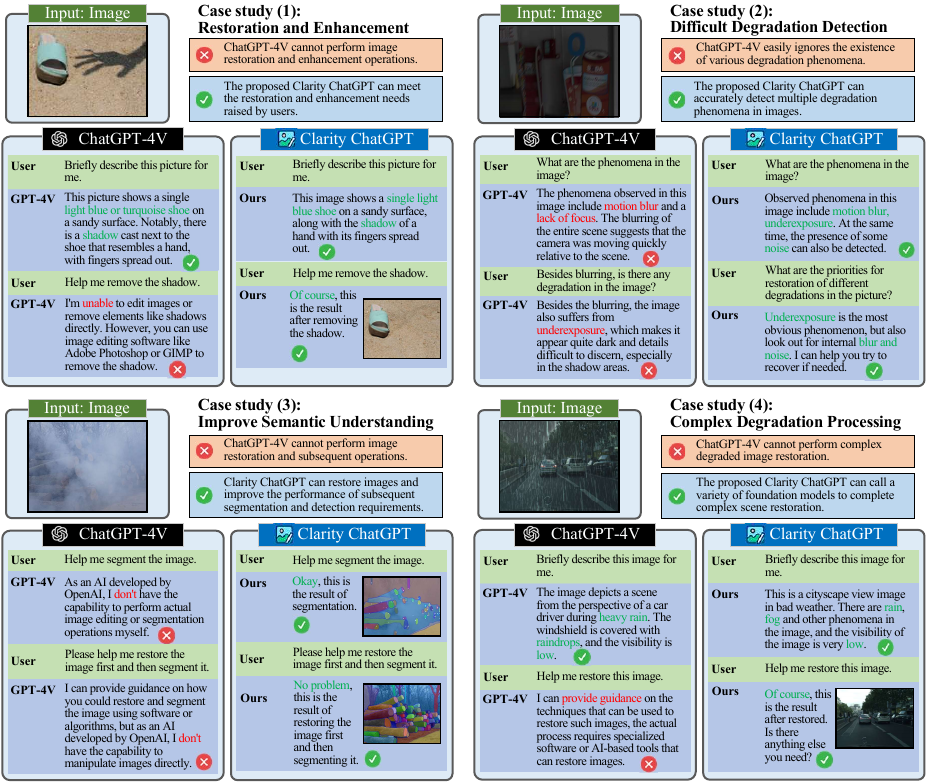}
    \vspace{-6mm}
	\caption{Case Study between ChatGPT-4V \cite{GPT-4} and proposed Clarity ChatGPT.}
	\label{fig:8}
    \vspace{-4mm}
\end{figure*}

We present a comprehensive case study to show the capabilities of Clarity ChatGPT. We exhibit its potential through various scenarios that showcase its proficiency in image restoration, enhancement, segmentation and detection, and addressing specific types of degradation in Figure \ref{fig:8}.

$\bullet$ \textbf{Case Study 1: Restoration and Enhancement.} Clarity ChatGPT demonstrates advanced image restoration, removing unwanted shadows with precision, notably improving upon ChatGPT-4V's limited editing capabilities, and delivering results that align closely with user intentions.

$\bullet$ \textbf{Case Study 2: Difficult Type Detection.} Clarity ChatGPT effectively identifies and corrects multiple image degradations including motion blur, underexposure, and noise—challenges that ChatGPT-4V recognizes only partially, thus offering a more thorough resolution.

$\bullet$ \textbf{Case Study 3: Improved Semantic Understanding.} Clarity ChatGPT excels in both understanding the semantic context of user requests and performing complex image processing tasks such as segmentation and detection, significantly outperforming ChatGPT-4V's capabilities.

$\bullet$ \textbf{Case Study 4: Complex Degradation Processing.} Clarity ChatGPT skillfully restores images with complex degradations, such as a cityscape in adverse weather conditions, utilizing a range of models for an enhanced restoration process that surpasses the capabilities of ChatGPT-4V.

Each case illustrates Clarity ChatGPT's advanced understanding and processing capabilities in the realm of image manipulation. By incorporating user feedback and leveraging state-of-the-art image processing algorithms, Clarity ChatGPT demonstrates a significant advancement over its predecessors, making it an invaluable tool for users requiring complex IRE tasks. More examples can be found in the \textit{supplementary material}.

\section{Conclusion}\label{Conclusion}
In this paper, we innovatively propose Clarity ChatGPT, a breakthrough system that introduces image-text interaction mechanism into the IRE domain. The core goal of Clarity ChatGPT is to provide dynamic, user-centric solutions for IRE. It firstly uses fine-tuned CLIP to accurately detect the degradation type of user input images, adaptively calls the relevant IRE model, and outputs the results. The system also provides region-specific optimization strategy and multiple results fusion technology to produce diverse results. At the same time, image quality evaluation and dialogue mechanisms can help users continuously interact and iteratively generate different results. Overall, Clarity ChatGPT is an effective attempt that not only enhances the adaptability and interactivity of the IRE but also demonstrates the great potential of combining natural language models with IRE.

\section{Future Work}\label{Future}
In our future work, we aim to enhance the performance and user experience of the Clarity ChatGPT system through the following initiatives: (1) \textbf{Open Platform Development}: We will promote the sharing and collaborative building of Demos, inviting a broader developer community to integrate a diverse range of IRE algorithms. (2) \textbf{Enhance User Feedback Mechanism}: We hope to establish a feedback loop, encouraging users to evaluate the results of image processing. The system will optimize management strategies based on user evaluations and continuously expand and refine functionalities accordingly. (3) \textbf{Model Scoring System}: A scoring system for models will be implemented, assigning recommendation weights to different models based on an analysis of the correlation between user preferences and model performance, to personalize user experience. 

\section{Acknowledgments}
This work is supported by the National Natural Science Foundation of China (62072151, 72004174, 61932009, 62020106007, 62072246), Anhui Provincial Natural Science Fund for the Distinguished Young Scholars (2008085J30), Open Foundation of Yunnan Key Laboratory of Software Engineering (2023SE103), CCF-Baidu Open Fund, CAAI-Huawei MindSpore Open Fund and the Fundamental Research Funds for the Central Universities (JZ2023HGQA0472). Corresponding author: Zhao Zhang. 

{\small
	\bibliographystyle{ieee_fullname}
	\bibliography{egbib}
}

\end{document}